\pdfoutput=1

\documentclass[11pt]{article}
\usepackage[final]{acl} 

\usepackage{times}
\usepackage{latexsym}
\usepackage[T1]{fontenc}
\usepackage[utf8]{inputenc}
\usepackage{microtype}
\usepackage{inconsolata}
\usepackage{graphicx}
\usepackage{booktabs}
\usepackage{amsmath}
\usepackage{array}
\usepackage{xcolor}
\usepackage{float}
\usepackage{amssymb}
\usepackage{comment}
\usepackage{soul}

\usepackage{booktabs}  
\usepackage{tabularx}  
\usepackage{listings}  
\usepackage{array}     
\usepackage{comment}
\usepackage{enumitem}

\begin{document}

\title{SAFE-SQL: Self-Augmented In-Context Learning with Fine-grained Example Selection for Text-to-SQL }

\author{Jimin Lee$^{1}$, ~Ingeol Baek$^{1}$, ~Byeongjeong Kim$^{1}$, ~Hyunkyung Bae$^{2}$, ~Hwanhee Lee$^{1}$\thanks{Corresponding author.}\\
$^{1}$Department of Artificial Intelligence, Chung-Ang University, $^{2}$New York University \\
\texttt{\{ljm1690, ingeolbaek, michael97k, hwanheelee\}@cau.ac.kr}, \texttt{hb3135@nyu.edu}
}

\maketitle

\begin{abstract}
Text-to-SQL aims to convert natural language questions into executable SQL queries. While previous approaches, such as skeleton-masked selection, have demonstrated strong performance by retrieving similar training examples to guide large language models (LLMs), they struggle in real-world scenarios where such examples are unavailable. 
To overcome this limitation, we propose \textbf{S}elf-\textbf{A}ugmentation in-context learning with \textbf{F}ine-grained \textbf{E}xample selection for Text-to-\textbf{SQL} (SAFE-SQL), a novel unsupervised framework that enhances SQL generation by generating and intelligently filtering self-augmented examples. 
SAFE-SQL leverages an LLM to generate diverse Text-to-SQL examples, which are then filtered by a novel fine-grained mechanism using criteria for semantic similarity, structural alignment, and reasoning path quality to curate high-quality in-context learning examples. Leveraging these carefully selected self-generated examples, SAFE-SQL significantly surpasses previous zero-shot and few-shot Text-to-SQL frameworks, achieving superior execution accuracy. Notably, our approach demonstrates substantial performance gains in challenging extra hard and unseen scenarios, where conventional methods often struggle.

\end{abstract}
\section{Introduction}

Text-to-SQL generation converts questions into SQL queries that help users access information in databases. 
Traditional approaches on Text-to-SQL rely on hand-crafted rules or simple pattern matching to generate SQL queries. They often struggle with the ambiguity and context-dependence of natural language, making it challenging to accurately translate user intent into structured SQL commands~\cite{SQLsurvey,rulebased_limitation,Rule_based_constructing}. As the field progressed, more sophisticated approaches emerged, including skeleton-masked selection~\cite{dail_sql}, relying on retrieving similar examples from training data to guide query generation. 
\begin{figure}[t]
\centerline{\includegraphics[scale=0.26]{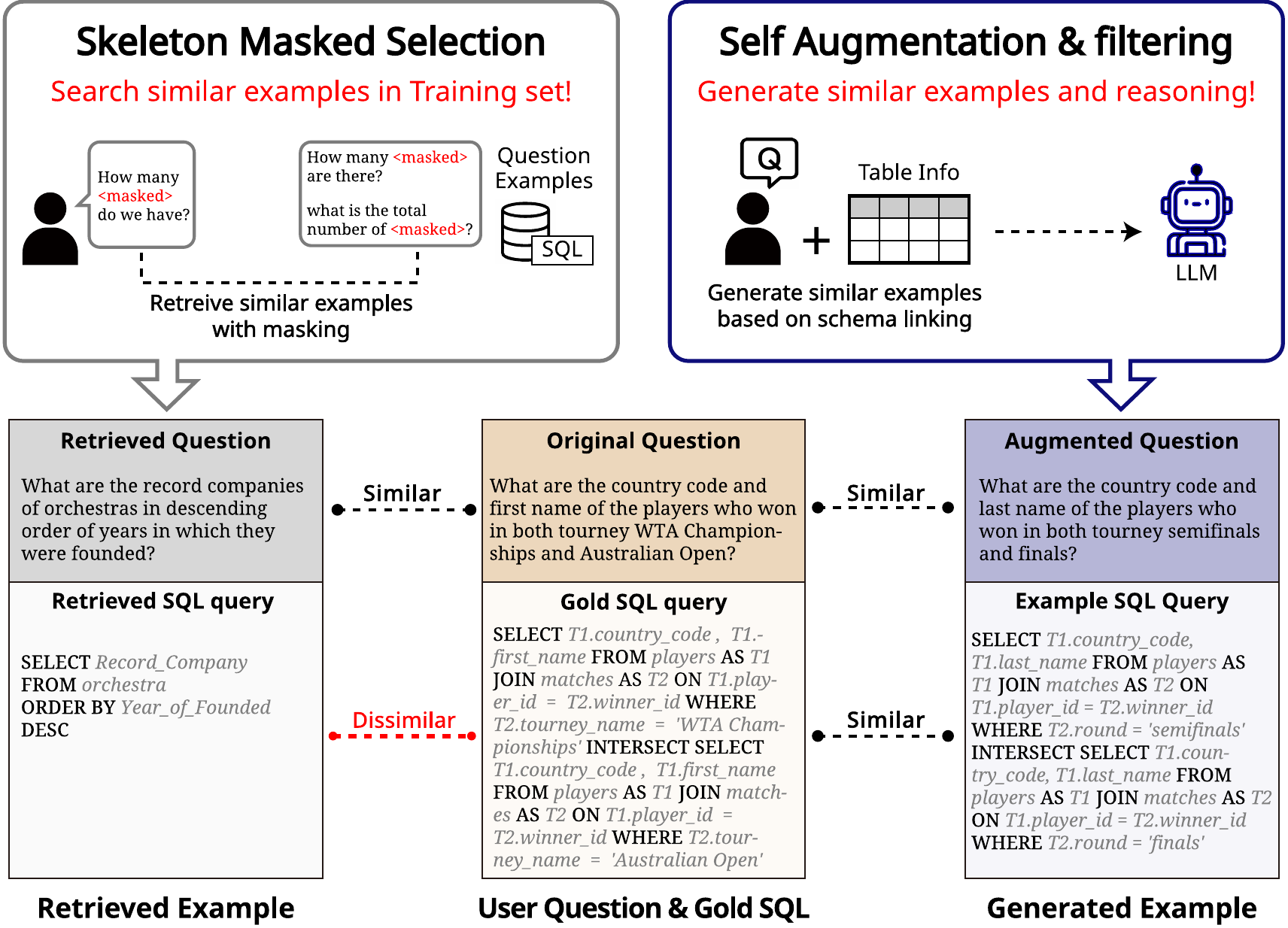}}
\caption{The example on the left shows a failure in retrieving relevant examples due to masked keywords, which results in superficially similar but actually unrelated questions being selected. In contrast, our self-augmented approach generates N-examples and filters them using 3 criteria, resulting in appropriate example retrieval.}
\label{fig:first_main}
\vspace{-5mm}
\end{figure}
However, these methods face significant challenges in real-world scenarios where similar examples are often unavailable in the training set~\cite{explore_limitation,Text_to_sql_survey} or unrelated examples are retrieved as shown in Figure~\ref{fig:first_main}. To overcome these problems, recent research has introduced methods to generate synthetic data. SQL-GEN, presents by~\cite{Dailect_gap_synthetic}, introduces dialect-specific synthetic data to resolve the diverse SQL dialect challenges in Text-to-SQL systems. Another important aspect of synthetic data generation is incorporating key relationships from the schema and employing schema-distance-weighted column sampling~\cite{Importance_of_synthesizeing}. However, these synthetic data generation methodologies predominantly require supervised fine-tuning, which demands substantial computational resources and time~\cite{synthesize}. 
Moreover, self-generated examples can introduce significant noise and inaccuracies that undermine the quality of in-context learning. Errors in synthetic SQL queries or flawed reasoning paths may lead to incorrect interpretations of database schemas~\cite{noise_text_to_sql}. As a result, relying on unfiltered self-generated examples for Text-to-SQL tasks can pose a risk of degrading overall model performance. 
Consequently, it is necessary to develop more efficient approaches that enhance the accuracy of Text-to-SQL while eliminating extra training costs and mitigating the adverse impacts of noisy self-generated examples by implementing a robust filtering mechanism.

In this paper, we propose SAFE-SQL, a novel approach that fully exploits the generative power of large language models (LLMs) to create high-quality synthetic examples in an unsupervised manner. 
SAFE-SQL enhances its inference capabilities without additional fine-tuning through four key steps: \textbf{\textit{(1) Schema Linking}}:
Analyzing SQL test questions, database tables, and foreign keys to map relationships between queries and database structures
\textbf{\textit{(2) Example Generation}}: 
Generating N-question-SQL query-reasoning path triplets per input using schema-linked information with LLMs
\textbf{\textit{(3) Threshold-based example selection}}: 
Filtering generated examples using specifically designed relevance criteria based on semantic similarity, Structural alignment, and reasoning path validity, retaining only those scoring above a specific threshold to ensure high quality and relevance for in-context learning examples and \textbf{\textit{(4) Final SQL Inference}}: 
Leveraging the curated examples, this step utilizes in-context learning to enhance the performance of large language models. This approach benefits from carefully selected examples that align with the natural language question and database schema, ensuring accurate and efficient SQL generation.

By relying on LLM-generated and filtered examples, SAFE-SQL significantly improves robustness and accuracy, particularly in complex or unseen scenarios where retrieval-based approaches struggle. Our approach eliminates the need for additional model training while achieving superior performance in Text-to-SQL tasks.
Our contributions can be listed as follows:
\begin{itemize}[leftmargin=10pt, labelindent=0pt]
    \item We propose SAFE-SQL, a fully unsupervised approach that leverages LLMs to generate synthetic examples. 
    \item Our method leverages schema linking to dynamically adapt examples, boosting the performance of Text-to-SQL in complex scenarios.
    \item We introduce a structured filtering mechanism that selects high-quality question-SQL pairs based on semantic similarity, structural alignment, and reasoning path validation.

\end{itemize}

\section{Related Work}

\begin{figure*}[!ht]
\centerline{\includegraphics[scale=0.3]{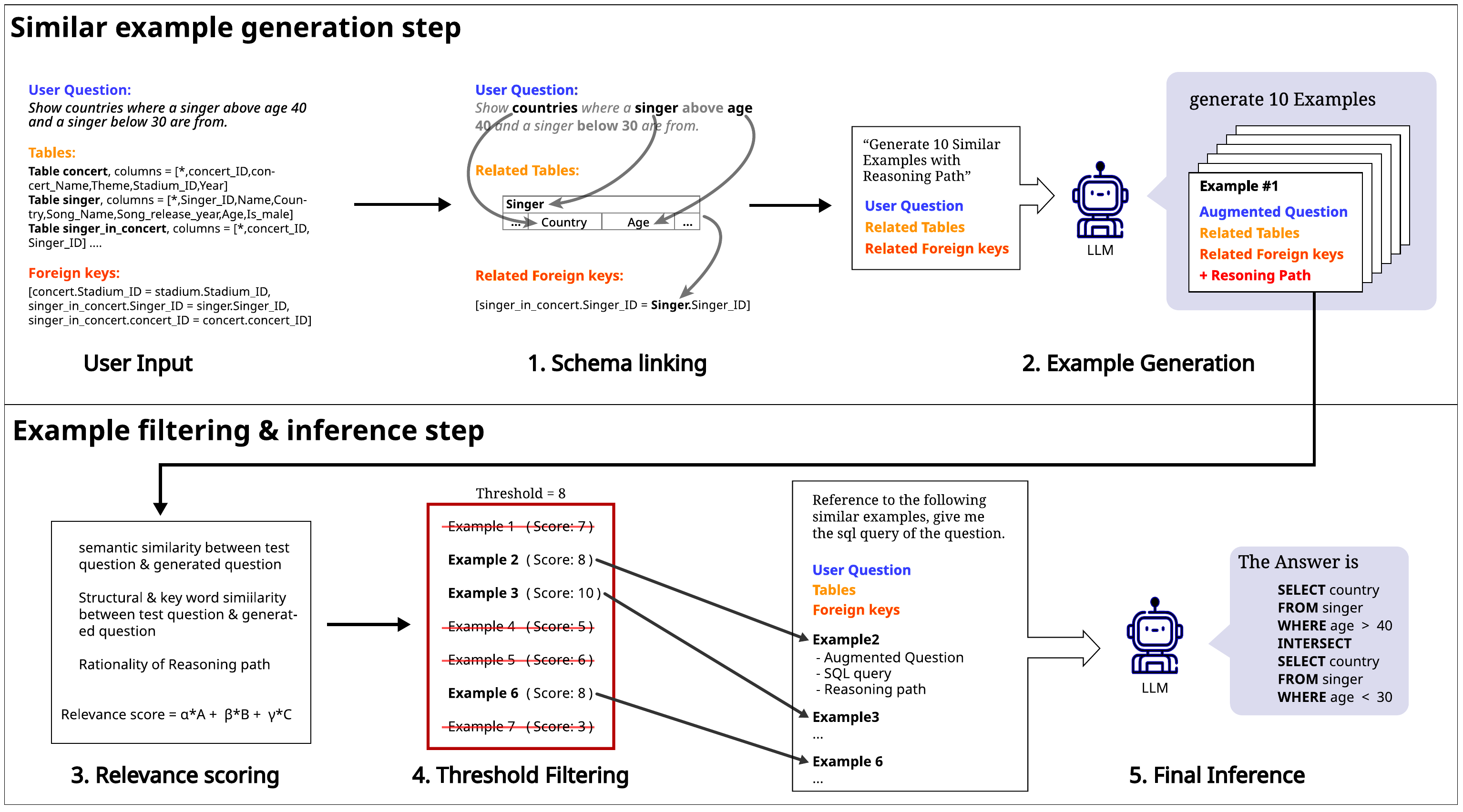}}
\caption{Overall flow of our proposed SAFE-SQL.}
\label{fig:main}
\end{figure*}


\paragraph{Structural and Semantic Information for Text-to-SQL}
Advances in Text-to-SQL have increasingly emphasized the importance of effectively utilizing structural and semantic information derived from the database schema. RAT-SQL~\cite{rat_sql} introduces relation-aware transformer architectures capable of encoding both the natural language question and the complex structure of the database schema, leading to improved schema linking performance. Concurrently, PICARD~\cite{picard} demonstrates that leveraging constrained decoding with step-by-step execution during generation can reduce the likelihood of producing invalid SQL queries. Rather than treating SQL generation as a pure sequence prediction task, PICARD executes partial SQL statements during generation, thus enforcing syntactic and semantic correctness. Both of these approaches highlight the importance of integrating structural and semantic information to generate correct SQL queries. Building on this line of research, our work explores how self-augmented examples can effectively incorporate structurally and semantically relevant information for in-context learning, particularly in scenarios where large annotated datasets are unavailable.


\paragraph{In-context Learning with Example Augmentation and Filtering}
As LLMs have demonstrated strong performance in in-context learning settings, recent work has focused on improving the effectiveness of demonstrations through better example augmentation and selection~\cite{coling_aug}. Self-Instruct~\cite{self_instruct} introduces a framework for generating instruction-tuning data by prompting the model to synthesize and filter examples, leveraging synthetic supervision. 
Further studies on demonstration selection for in-context learning~\cite{demonstrate_selection} have systematically studied strategies for selecting effective in-context examples, including similarity-based retrieval and clustering methods. 
Integrating self-generated prompts with explicit reasoning chains has also been shown to significantly improve in-context learning outcomes by guiding the model's thought process~\cite{reasoning, cot_reasoning}. These studies collectively underscore the impact of demonstration quality on LLM performance and highlight the potential of intelligently curating examples. 
Focusing on the Text-to-SQL task, our work distinguishes itself by generating synthetic examples and filtering them using a novel fine-grained mechanism that considers semantic similarity, structural similarity, and reasoning path quality. This allows us to generate multiple candidate examples and select the most effective through this tailored fine-grained filtering process.

\section{Fine-grained Self-Augmentation for Text-to-SQL}
We propose SAFE-SQL, a framework that automatically generates high-quality examples for in-context learning in Text-to-SQL tasks. 
Unlike traditional methods that rely on retrieving similar questions or using predefined templates, SAFE-SQL uses LLMs to create synthetic examples tailored to the given database schema.
These examples are then filtered based on their \textit{semantic similarity}, \textit{structural alignment}, and the \textit{quality of reasoning paths}.
Finally, we predict the final SQL query for the test input using the self-generated examples via in-context learning.

\subsection{Schema Linking}
The first step in SAFE-SQL is schema linking, which identifies and extracts relevant schema elements from the database to reduce noise and improve performance in Text-to-SQL tasks~\cite{robust_schema_linking}.
As shown in Figure~\ref{fig:main}, the schema linking step involves analyzing the test question to detect keywords and phrases that correspond to schema elements such as tables, columns, rows, and foreign keys within the database schema.
This mapping narrows the focus to the most pertinent parts of the schema and provides the necessary context for generating relevant examples that are both meaningful and grounded in the database structure. 
\subsection{Example Generation}
Using the information obtained from schema linking, the LLM generates a pool of multiple synthetic examples for each test question. As illustrated in Figure~\ref{fig:main}, for each test question, we generate ten examples—each comprising a similar question, its corresponding SQL query, and a detailed reasoning path. The generated SQL questions maintain structural similarity while varying elements such as numerical values, table names, and key attributes. This ensures that the generated examples remain relevant while encouraging the model to generalize beyond surface-level patterns. By observing these modified instances, the model can infer the correct SQL query even when face with unseen but structurally similar questions. In particular, the reasoning path outlines the logical steps required to derive the correct SQL query result, providing a comprehensive explanation of the query execution process. We provide the full prompt used for LLMs in Appendix~\ref{appen:example_prompt}.

\subsection{Relevance Scoring}
After generating a set of synthetic examples, SAFE-SQL employs a crucial evaluation process rooted in novel fine-grained example selection to determine the relevance of each generated example to the test question. 
This fine-grained selection process is integral to our method, ensuring that only high-quality, contextually appropriate examples are used for in-context learning, moving beyond simple retrieval to curate examples truly relevant and beneficial for the Text-to-SQL task. 

To achieve this, we assign composite relevance score \( Rel\) on a scale from 0 to 10 to each example \( e \), which is calculated as follows: 
\begin{equation}\resizebox{0.42\textwidth}{!}{
$ Rel = \alpha \cdot S(Q_e, Q_t) + \beta \cdot A(Q_e, Q_t) + \gamma \cdot R$}
\end{equation}

Here, $Q_t$ represents the test question, $Q_e$ denotes the generated example question. The coefficients $\alpha$, $\beta$, and $\gamma$ are weighting factors that sum to 1, allowing for adjustment of the relative importance of each component in the fine-grained selection score.
The three components are defined as follows:

\begin{itemize}
    \item \textbf{Semantic Similarity \( S(Q_e, Q_t) \):} assesses if the generated question preserves the underlying meaning and intent to ensure the example aligns with the user's core query objective.
    \item \textbf{Structural Alignment \( A(Q_e, Q_t) \):} evaluates structural correspondence based on key database elements and their relationships, which is important for mapping natural language to a similar database structure and operations.
    \item \textbf{Reasoning Path Quality \( R \):} evaluates the alignment of the example's logical derivation steps and database operations (e.g., filtering, aggregation, joins, subqueries) with the test question's required logic.
\end{itemize}
We utilize LLMs to compute the score for each of the three components based on our specifically designed instructions and criteria.
Specifically, the LLM applies a predefined, multi-point scoring rubric (detailed in Appendix~\ref{appen:filtering_examples}) to assign a quantitative score (0-10) for each criterion.
This process allows for a nuanced assessment of the degree of alignment between the generated example and the test question along each dimension, moving beyond simple binary or qualitative judgments.
By carefully evaluating these three factors through our fine-grained example selection process, SAFE-SQL ensures that the selected examples are highly relevant and informative, contributing to more accurate and effective SQL query generation.


\subsection{Threshold Selection}
To further ensure quality, SAFE-SQL retains only those examples with a relevance score above a predefined threshold \( \theta \). Formally, the set of selected examples is defined as:
\begin{equation}\resizebox{0.30\textwidth}{!}{
$E_{\text{selected}} = \{ e \in E \mid Rel \geq \theta \}$}
\end{equation}
where \( E \) represents all generated examples. This thresholding step filters out low-quality examples and ensures that only the most informative and contextually appropriate examples are used in the final inference. The threshold is set to 8, as Figure~\ref{tab:diff_thres} demonstrates that this value provides an optimal balance between preserving high-quality examples and maintaining sufficient diversity for robust SQL generation.

\subsection{Final Inference}
In the final stage, the high-quality examples generated in previous steps are combined with the test question to construct a comprehensive prompt for the LLM. These examples, enriched with filtered questions, corresponding SQL queries, and detailed reasoning paths, guide the LLM in generating the final SQL query. By integrating schema linking, synthetic example generation, relevance scoring, and threshold-based filtering, SAFE-SQL produces SQL queries that are both syntactically correct and semantically aligned with the intended database operations, while also providing an interpretable reasoning process.

\section{Experiment}
\subsection{Experimental Setup}
For our experiments, we employ six models for comparison purposes: GPT-4o~\cite{hurst2024gpt}, GPT-4o-mini~\cite{hurst2024gpt}, GPT-4~\cite{achiam2023gpt}, Llama-3.1-70B-Instruct~\cite{dubey2024llama}, Llama3.3-70B-Instruct~\cite{dubey2024llama}, Qwen2.5-72B~\cite{yang2024qwen2},  Gemma3-12b~\cite{gemma3}, and Gemma3-27b~\cite{gemma3}. The evaluation is conducted on the Spider dev dataset~\cite{spider} and the Bird dev dataset~\cite{bird_dataset}, which are widely used benchmarks for Text-to-SQL systems. The Spider dev set contains 7,000 training samples covering 166 databases in various domains and 1,034 evaluation samples from 20 databases, comprised of four difficulty levels. BIRD is a large cross-domain Text-to-SQL dataset with 12,751 question-SQL pairs across 95 databases. Since the test sets of both the Spider and BIRD datasets are only accessible through specific evaluation servers, we conduct our evaluation using their respective development sets.


\subsection{Baselines}
We use the following baseline Text-to-SQL methods: \textbf{Supervised fine tuning}, which fine-tunes an open source model, \textbf{Zero-shot}, which infers without examples, \textbf{Few-shot}, which infers with few examples. \textbf{Synthesizing Text-to-SQL data from weak and strong LLMs}~\cite{synthesize} utilizes preference learning from the weak data from small LLMs and strong data from LLMs. \textbf{SQL-PaLM}~\cite{palm-sql} introduces synthetic data augmentation to fine-tune open source models. \textbf{Din SQL}~\cite{din} breaking down the task into smaller sub-tasks, allowing large language models to improve their reasoning process through self-correction iteratively. \textbf{C3-SQL}~\cite{c3_zeroshot} comprises \textit{clear prompting}, \textit{calibration with hints}, and \textit{consistent output}, which systematically addresses model input, bias, and output to enhance performance using the zero-shot prompt. \textbf{Dail-SQL}~\cite{dail_sql} introduces effective few-shot learning, significantly reducing the number of tokens required per question. \textbf{ACT-SQL}~\cite{act_sql} enhances Text-to-SQL performance by automatically generating chain-of-thought exemplars, eliminating the need for manual labeling. \textbf{PTD-SQL}~\cite{ptd} categorizes queries into subproblems and focuses on targeted drilling to improve LLMs' reasoning capabilities. 

\begin{table}[t]
\setlength{\tabcolsep}{3pt} 
\centering
\resizebox{\columnwidth}{!}{
\small
\begin{tabular}{llcccccc}
\toprule
\textbf{Method} & \textbf{Model} & \textbf{Easy} & \textbf{Medium} & \textbf{Hard} & \textbf{Extra} & \textbf{All} & \textbf{Time}\\ 
\midrule
\multicolumn{7}{c}{\textbf{Supervised Fine-Tuning (SFT)}} \\
\midrule
SYN-SQL    & Sense-13B           & \textbf{95.2} & 88.6 & 75.9 & 60.3 & 83.5 & - \\  
SQL-PaLM   & PaLM-2               & 93.5 & 84.8 & 62.6 & 48.2 & 77.3 & - \\  
\midrule
\multicolumn{7}{c}{\textbf{Zero-shot Methods}} \\
\midrule
Baseline   & GPT-4                & 84.3 & 73.1 & 65.8 & 40.3 & 69.1 & 1.28\\   
Baseline   & GPT-4o                & 87.2 & 77.2 & 68.4 & 48.7 & 73.4 & 0.93\\  
Baseline   & GPT-4o-mini          & 84.8 & 75.6 & 67.0 & 46.1 & 71.5  & 1.07\\    
C3-SQL    & GPT-4                & 90.2 & 82.8 & 77.3 & 64.3 & 80.6 & 19.34\\  
\midrule
\multicolumn{7}{c}{\textbf{Few-shot Methods}} \\
\midrule
DIN-SQL    & GPT-4                & 91.1 & 79.8 & 64.9 & 43.4 & 74.2 & 4.37\\ 
DAIL-SQL   & GPT-4                & 91.9 & \textbf{90.1} & 75.2 & 63.8 & 83.6 & 16.79\\ 
ACT-SQL    & GPT-4                & 91.1 & 79.4 & 67.2 & 44.0 & 74.5 & 4.55\\
PTD-SQL    & GPT-4                & \underline{94.8} & 88.8 & 85.1 & 64.5 & 85.7 & 7.89\\ 
DEA-SQL    & GPT-4                & 88.7 & \underline{89.5} & 85.6 & 70.5 & 85.6 & 8.69\\ 
\midrule
\multicolumn{7}{c}{\textbf{Self-augmented In-Context Learning}} \\
\midrule
SAFE-SQL    & GPT-4                & 93.2 & 88.9 & 85.8 & 74.7 & 86.8 & 21.41\\ 
SAFE-SQL    & GPT-4o                & 93.4 & 89.3 & \textbf{88.4} & 75.8 & 87.9 & 14.92\\
SAFE-SQL    & GPT-4o-mini         & 93.6  & 87.5 & 86.1 & 75.2 & 87.4 & 15.33\\
SAFE-SQL    & Llama3.1-70B-Instruct        & 90.4  & 88.2 & \underline{86.2} & \underline{78.2} & 86.8 & 23.52\\
SAFE-SQL    & Llama3.3-70B-Instruct         & 92.0  & 80.5 & 81.0 & 62.9 & 80.5 & 22.46\\
SAFE-SQL    & Qwen2.5-72B         & 87.6  & 74.5 & 77.0 & 52.4 & 74.5 & 28.51\\
SAFE-SQL    & Gemma3-12B        & 92.4  & 90.7 & 85.1 & \textbf{78.8} & \underline{88.2} & 13.28\\
SAFE-SQL    & Gemma3-27B        & 93.6  & 89.8 & 87.4 & \textbf{78.8} & \textbf{88.5} & 14.26\\
\bottomrule
\end{tabular}
}
\caption{Execution accuracy across difficulty levels on the Spider development set. The highest score per row is in bold, and the second highest is underlined.}
\label{tab:sql_comparison}
\end{table}

\subsection{Evaluation Metrics}
We use Execution Accuracy (EX) and Exact Match (EM) to evaluate the performance of our model. EX measures whether the SQL query generated by the model produces the same results as the ground truth query when executed on a database. Exact Match (EM), on the other hand, assesses whether the predicted SQL query exactly matches the ground truth query in its structure and syntax. By combining these two metrics, we ensure a comprehensive evaluation of both the correctness and execution reliability of the generated SQL queries.

\begin{table}[t]
\centering
\resizebox{0.85\linewidth}{!}{
\small
\begin{tabular}{lcc}

\toprule
\textbf{Method} & \textbf{Model} & \textbf{Execution Accuracy} \\ 
\midrule
\multicolumn{3}{c}{\textbf{Supervised Fine-Tuning (SFT)}} \\
\midrule
Syn-SQL    & Sense13B         & \underline{63.4} \\  
SQL-Palm   & Palm             & 53.6 \\  
\midrule
\multicolumn{3}{c}{\textbf{Zero-shot Methods}} \\
\midrule
Baseline   & GPT-4            & 49.2 \\   
Baseline   & GPT-4o            & 51.8 \\  
Baseline   & GPT-4o-mini       & 51.2 \\    
C3-SQL     & GPT-4            & 53.8 \\  
\midrule
\multicolumn{3}{c}{\textbf{Few-shot Methods}} \\
\midrule
Din-SQL    & GPT-4            & 55.9 \\ 
Dail-SQL   & GPT-4            & 55.4 \\ 
ACT-SQL    & GPT-4            & 52.8 \\
PTD-SQL    & GPT-4            & 57.0 \\ 
DEA-SQL    & GPT-4            & 52.4 \\ 
\midrule
\multicolumn{3}{c}{\textbf{Self-augmented In-Context Learning}} \\
\midrule
SAFE-SQL   & GPT-4            & 58.9 \\ 
SAFE-SQL   & GPT-4o           & \textbf{63.5} \\
SAFE-SQL   & GPT-4o-mini      & 62.1 \\
SAFE-SQL   & Llama3.1-70B-Instruct            & 60.9 \\ 
SAFE-SQL   & Llama3.3-70B-Instruct           & 61.2 \\
SAFE-SQL   & Qwen2.5-72B      & 56.2 \\
SAFE-SQL   & Gemma3-12B            & 60.8 \\ 
SAFE-SQL   & Gemma3-27B            & 61.5 \\ 
\bottomrule
\end{tabular}}
\caption{Execution accuracy on Bird dataset.}
\label{tab:bird}
\end{table} 


\subsection{Performance Comparison}
\paragraph{Spider Dataset}
We analyze the performance of SAFE-SQL across different SQL difficulty levels and compare it with zero-shot, few-shot prompting methods, and supervised fine-tuning approaches. The results, presented in Table~\ref{tab:sql_comparison}, demonstrate that SAFE-SQL achieves overall superior performance, with particularly strong improvements in hard and extra hard categories. Few-shot methods exhibit higher accuracy in Easy and Medium categories, which can be attributed to skeleton-masked selection which retrieves answers directly from the training set, leading to an inflated performance in simpler queries. SAFE-SQL excels in hard and extra hard categories, achieving significantly higher EX. This improvement is notably influenced by the inclusion of reasoning paths, which provide explicit guidance in SQL generation and enhance the model’s ability to construct complex queries, as well as the filtering of misleading examples, which reduces potential confusion and prevents error propagation. These multiple factors play a crucial role in enabling the model to generate more accurate and structurally sound SQL queries, especially in challenging scenarios where other approaches struggle. Notably, SAFE-SQL using open-source models such as Gemma3-27B outperforms high-cost methods based on GPT-4, highlighting its cost-effectiveness and strong capability.

\paragraph{Bird Dataset}
We also conduct experiments on the Bird Dev dataset in addition to the Spider dataset. Similar to Spider, SAFE-SQL consistently outperforms zero-shot and few-shot methods, achieving 63.5\% execution accuracy with GPT-4o, which is even higher than Syn-SQL (63.4\%), a supervised fine-tuning approach. This highlights the effectiveness of our SAFE-SQL using a self-augmented in-context learning method.

\begin{table}[t]
    \centering
    \small
    \begin{tabular}{lcc}
        \toprule
        Models & EX  & EM \\
        \midrule
        SAFE-SQL - GPT-4o & 87.9 & 78.3\\ 
        w/o Reasoning path & 84.4 (-3.5) & 73.6(-4.7) \\
        w/o Relevance filtering & 82.1 (-5.8) & 68.5(-9.7) \\
        w/o Schema linking & 80.4 (-7.5)  & 65.1(-13.2)\\
        w/o Similar examples & 77.1 (-10.8) &  61.9(-16.4)\\
        \bottomrule
    \end{tabular}
    \caption{Ablation study results for SAFE-SQL, where removing each component leads to a performance drop.}
    \label{tab:ablation}
\end{table}

\subsection{Ablation Study}
To assess the contribution of each key component in our model, we conduct an ablation study by systematically removing four critical modules: \textbf{Reasoning Path}, \textbf{Relevance Score}, \textbf{Schema Linking}, and \textbf{Similar Examples}.  We evaluate the resulting impact on performance using EX shown in Table~\ref{tab:ablation}. Our findings indicate that each component plays a crucial role in the model’s effectiveness. Removing the Reasoning Path leads to a 3.5-point drop in EX, highlighting its importance in guiding the model toward generating accurate SQL queries. The absence of the Relevance Score resulted in a 5.8-point decrease in EX, underscoring its contribution to overall performance. Eliminating Schema Linking causes a 7.5-point drop in EX, which demonstrates its critical role in similar example construction. Overall, each of the four components—Reasoning Path, Relevance Score, Schema Linking, and Similar Examples—is essential for achieving optimal performance in SQL generation, empirically validating our architectural and design choices. 
\subsection{Analysis}

\begin{figure*}[t]
\centerline{\includegraphics[scale=0.5]{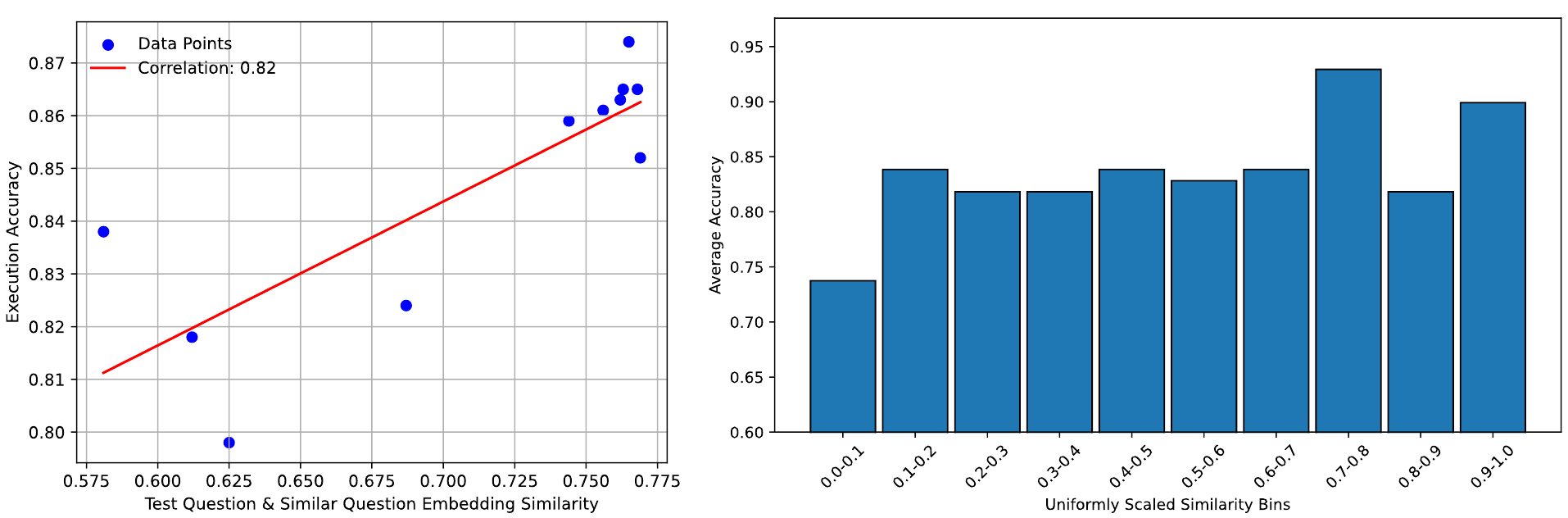}}

\caption{(Left) Correlation between question embedding similarity and average EX, (Right) Average EX across embedding similarity bins}
\vspace{-3mm}
\label{fig:corr_bin}
\end{figure*}

\paragraph{Inference Time per Query on Spider Dev Set}
As shown in Table~\ref{tab:sql_comparison}, we compare the inference time of SAFE-SQL with other methods. While baseline methods achieve faster inference via simple zero-shot prompts, they show lower accuracy. Few-shot methods are faster than SAFE-SQL but still underperform in execution accuracy. In contrast, SAFE-SQL leverages example augmentation and filtering process, achieving higher performance with a modest increase in inference time. Despite requiring three LLM calls, SAFE-SQL demonstrates strong zero-shot capabilities without relying on a training set, making the trade-off in latency worthwhile.

\begin{table}[h]
    \centering
    \small
    \begin{tabular}{lccc}
        \toprule
        Score & cos $\theta$ &\# of Generated EX & \%  Filtered EX \\
        \midrule
        \textbf{$\geq 0$} &0.581& 10340 & 0 \% \\ 
        \textbf{$\geq 2$} &0.625& 10185  & 1.50\% (-155) \\
        \textbf{$\geq 4$} &0.744& 9883 & 4.41\% (-457)  \\
        \textbf{$\geq 6$} &0.762 & 9378 & 9.30\% (-962)  \\
        \textbf{$\geq 8$}&0.765& 8606 & 16.76\% (-1734)\\
        \textbf{$\geq 10$} &0.769& 6795 & 34.28\% (-3545)  \\
        \bottomrule
    \end{tabular}
\caption{Summary of data generation, filtering results, and embedding similarity analysis by score.}
    \label{tab:number_of_generated}
\end{table}


\paragraph{Number of Generated and Filtered Examples per Score, along with an Embedding Similarity Analysis of the Filtered Examples}
For each test question in the Spider dev set, 10 examples are generated, resulting in a total of 10,340 examples. The quality of these examples is assessed using a relevance score ranging from 0 to 10. As shown in Table~\ref{tab:number_of_generated}, the 65.71\% of examples are assigned a score of 10, while the 0.59\% of examples are received a score of 0. This trend suggests that the LLM tends to assign high relevance to its own generated examples. The similarity is computed using cosine similarity, where higher scores indicate greater semantic alignment between the test questions and the retained examples. As the filtering threshold increases, the embedding similarity also increases, suggesting that higher-relevance examples exhibit stronger semantic consistency with the test questions. However, we also observe that overly strict filtering—selecting only examples with a perfect score of 10—leads to a decline in performance. This drop occurs because an excessively high threshold significantly reduces the number of available examples, limiting the diversity.

\paragraph{Effect of Question Embedding Similarity on Execution Accuracy.}
In Figure~\ref{fig:corr_bin}, the left graph illustrates the correlation between embedding similarity and EX. Each point represents one of the 11 data points obtained by filtering examples based on different threshold scores (0 to 10). The data points follow an upward trend, suggesting that higher similarity tends to result in better EX. The red line indicates the overall correlation, with a coefficient of 0.82, showing a relatively strong positive relationship. Building on this analysis, the right graph provides a more fine-grained view by examining the execution accuracy of individual generated examples based on their embedding similarity with test questions. The x-axis represents the normalized similarity between the test question and the generated question, and the y-axis indicates EX. The results show that EX is lowest in the 0.0-0.1 similarity range, suggesting that examples with very low similarity to test questions tend to be less useful. As similarity increases, EX generally improves, peaking in the 0.7-0.8 range. This suggests that examples with a moderate to high similarity to test questions are more effective in generating executable SQL queries. However, accuracy drops slightly in the 0.8-0.9 range before rising again in the 0.9-1.0 range. This indicates that excessively high similarity can reduce diversity, potentially limiting the model’s generalization ability.

\begin{figure}[t]
\centerline{\includegraphics[scale=0.36]{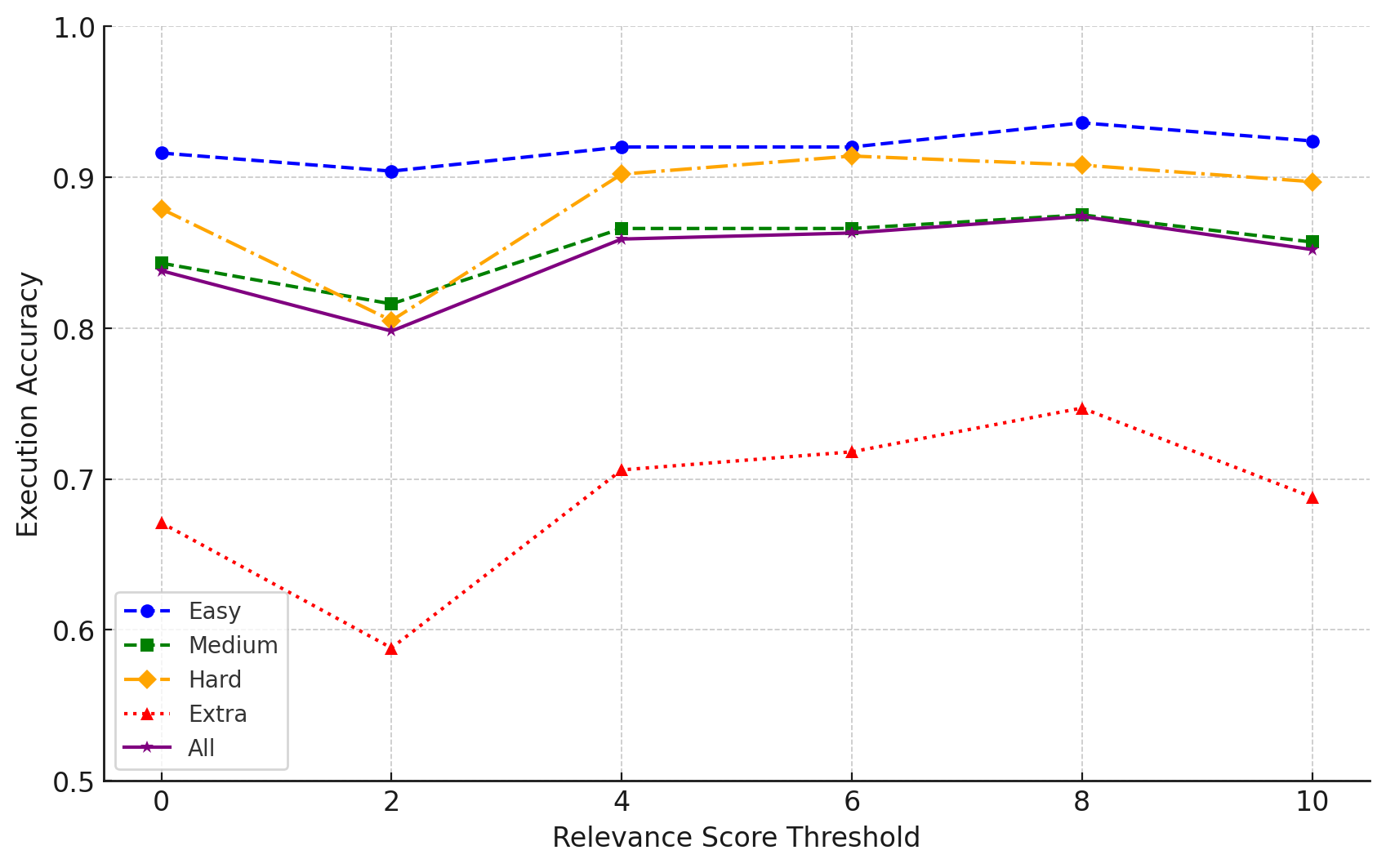}}
\caption{Performance of GPT-4o at different relevance score thresholds.}
\label{tab:diff_thres}
\end{figure}

\begin{table*}[t]
    \centering
    \scriptsize
    \renewcommand{\arraystretch}{1.2}
    \resizebox{1.99\columnwidth}{!}{
    \begin{tabularx}{\textwidth}{X X X >{\raggedright\arraybackslash}p{0.3\textwidth} X}
        \toprule
        \textbf{GOLD Question} & \textbf{GOLD SQL Query} & \textbf{Generated Question} & \textbf{Reasoning Path} & \textbf{Relevance Score} \\
        \midrule
        \hl{Question1:} What are the names, countries, and ages for every singer in descending order of age? & 
        \texttt{SELECT name, country, age FROM singer ORDER BY age DESC} & 
        \sethlcolor{lime!50}\hl{What are the names, ages, and countries of all singers from a specific country, sorted by age in descending order?} & 
        \sethlcolor{violet!20}\hl{1. Identify the desired columns: name, age, and country. \\
        2. Specify the table: singer. \\
        3. Sort the results by age in descending order.} &  
        Semantic similarity = 10 
        Structural Similarity = 10  
        Reasoning path = 10  
        Relevance score = (10+10+10)/3 = 10
                \\ 
        \midrule
        \hl{Question2:} Return the names and template ids for documents that contain the letter w in their description. & 
        \texttt{SELECT document\_name, template\_id FROM Documents WHERE Document\_Description LIKE "\%w\%"} & 
        \sethlcolor{lime!50}\hl{Retrieve the titles and category IDs of articles whose summaries contain the word "data".} &
        \sethlcolor{violet!20}\hl{1. Identify the necessary columns: extract title and category\_id from the Articles table. \\
         2. Locate the relevant table: use the Articles table as it contains the required data. \\
        3. Define the filtering condition: apply a WHERE clause to check if the summary column contains the substring "data". \\
         4. Use the LIKE operator: employ LIKE '\%data\%' to search for any instance of "data" within the summary. \\
        5. Retrieve the results: return the title and category\_id values for all matching records.} & Semantic similarity = 7
                    Structural Similarity = 9
Reasoning path = 8
Relevance score = (7+9+8)/3 = 8
        \\
        \midrule
        \hl{Question3:} What is the number of car models that are produced by each maker and what is the id and full name of each maker? &  
        \texttt{SELECT Count(*), T2.FullName, T2.id FROM MODEL\_LIST AS T1 JOIN CAR\_MAKERS AS T2 ON T1.Maker = T2.id GROUP BY T2.id;} & 
        \sethlcolor{lime!50}\hl{List all employees who work in the IT department along with their employee ID and hire date.} & 
        \sethlcolor{violet!20}\hl{1. Identify required details: employee ID and hire date. \\
        2. Filter condition: find employees who work in IT. \\
        3. Retrieve data: select only emp\_id and hire\_date.} & 
        Semantic similarity = 6
        Structural Similarity = 3
        Reasoning path = 2
        Relevance score = (6+3+2)/3 = 3.67
  \\
        \bottomrule
    \end{tabularx}
    }
    \caption{Comparative examples of gold-questions and GPT-4o-generated SQL questions, including detailed reasoning paths and the process of computing relevance scores.}
    \label{tab:sql_examples}
\end{table*}

\begin{table}[t]
    \centering
    \small
    \resizebox{0.48\textwidth}{!}{
    \begin{tabular}{lcc||ccccc}
        \toprule
        \textbf{$\alpha$} & \textbf{$\beta$} &\textbf{$\gamma$}& \textbf{Easy}& \textbf{Medium}& \textbf{Hard} &\textbf{Extra}& \textbf{EX} \\
        \midrule
        0.33 & 0.33 & 0.33 & \textbf{93.4} & \textbf{89.3} & \underline{88.4} & \textbf{75.8} & \textbf{87.9} \\ 
        \midrule
        1 & 0 & 0 & 90.7& 84.2& 82.3& 68.3&  82.8 \\ 
        0 & 1 & 0 & 89.8& 85.6& 81.2& 69.2&  83.1 \\ 
        0 & 0 & 1 & 89.2& 85.1& 84.3& 71.7& 83.7  \\ 
        \midrule
        0.5 & 0.5 & 0& 91.2& 87.3& 82.5& 69.4& 84.4 \\ 
        0.5 & 0 & 0.5& 92.5& \underline{87.9}& 83.5& 70.3& 85.3 \\ 
        0 & 0.5 & 0.5& \underline{92.7}& 86.8& \textbf{88.5}& \underline{72.4}& \underline{86.1} \\ 
        \bottomrule
    \end{tabular}
    }
    \caption{Execution accuracy across difficulty levels under different weights: semantic similarity ($\alpha$), structural similarity ($\beta$), and reasoning path quality ($\gamma$).}
     \vspace{-3mm}
    \label{tab:filtering_score_ablation}
\end{table} 

\paragraph{Effect of Relevance Scoring Thresholds on Performance.}

To further evaluate the effectiveness of SAFE-SQL, we conduct a detailed case study using varying thresholds for the relevance scoring mechanism as shown in Figure~\ref{tab:diff_thres}.  The self-generated examples are filtered based on relevance scores, with thresholds ranging from 0 to 10. For each test question, the number of high-scoring examples varied due to the specific content and schema structure (e.g., some test questions have six examples with scores $\geq 8$, while others have three). The selected examples are then used during the final inference stage to generate SQL queries. The $\geq 8$ threshold consistently produces the best results, validating the robustness of SAFE-SQL’s relevance score filtering. The results demonstrate that selecting high-quality examples plays a critical role in guiding LLMs to generate accurate SQL queries, regardless of the underlying model.


\paragraph{Effect of Three Measuring Components on Performance.}

To assess the impact of the three measuring components—semantic similarity ($\alpha$), structural similarity ($\beta$), and reasoning path quality ($\gamma$)—on EX, we conduct experiments by varying their respective weightings. The results, presented in Table~\ref{tab:filtering_score_ablation}, highlight distinct performance trends across different difficulty levels. Notably, the exclusion of reasoning path quality leads to a drop in EX, particularly in the Hard and Extra Hard. This suggests that a well-structured reasoning path is crucial for handling complex queries, as it provides essential logical steps that bridge the gap between natural language understanding and SQL formulation. Conversely, semantic similarity and structural similarity have a greater influence on the Easy and Medium levels. This is because these queries tend to be relatively straightforward, meaning that having structurally similar SQL questions in the example set often provides sufficient guidance for generating correct queries. In these cases, direct pattern matching and schema alignment play a larger role. Overall, the findings demonstrate that a balanced combination of all three components is essential for optimizing performance across different levels of query complexity.



\subsection{Case Study}
As shown in Table~\ref{tab:sql_examples}, test questions from the Spider dev set alongside their generated similar examples, evaluated based on semantic similarity, structural similarity, and the reasoning path score, which together determine the relevance score. The first example achieves a perfect relevance score of 10, as the generated question closely aligns with the original in meaning, structure, and reasoning. The SQL formulation remains nearly identical, and the reasoning path explicitly details each step, ensuring full alignment. The second example receives a relevance score of 8, with semantic similarity of 7 due to minor differences in terminology ("documents" vs. "articles" and "letter 'w'" vs. "word 'data'"). However, its structural similarity remains high, as the SQL structure is nearly identical. The reasoning path score of 8 reflects a clear explanation of query formulation, though slightly less detailed than the first example. The third example has the lowest relevance score due to significant differences. The generated question shifts focus from counting car models to listing IT employees, resulting in semantic similarity of 6 and structural similarity of 3. These results emphasize the importance of fine-grained example selection due to the varing quality of generated examples.

\section{Conclusion}
We introduce SAFE-SQL, a novel unsupervised framework designed for Text-to-SQL. SAFE-SQL generates and filters high-quality self-augmented examples for in-context learning. Extensive experiments demonstrated that both the fine-grained example generation process and optimal threshold filtering contribute significantly to performance gains. Our method achieves state-of-the-art results, showing notable improvements over ablated versions and excelling particularly in challenging extra hard and unseen scenarios. 

\section*{Limitations}
While SAFE-SQL demonstrates strong performance in generating accurate and semantically valid SQL queries, there are a few limitations that should be addressed in future work. Although the model performs well on the tested datasets, its ability to generalize to highly diverse or domain-specific SQL tasks remains to be fully evaluated. The current framework also relies on large language models like GPT-4o, which may not be easily scalable to low-resource settings or environments with limited computational resources.  Handling edge cases and extremely complex queries, which might require deeper schema understanding and more sophisticated reasoning, is another challenge for the model. 

\section*{Ethics Statement}
While our approach enhances SQL generation without additional fine-tuning, it relies on LLMs, which may inherit biases from training data. We mitigate potential biases and inaccuracies through structured filtering and relevance scoring. Our study uses publicly available datasets, ensuring compliance with data privacy standards. We encourage responsible use of our method, particularly in applications requiring high accuracy and fairness.


\section*{Acknowledgement}
This work was supported by the Institute of Information \& Communications Technology Planning \& Evaluation (IITP) grant funded by the Korea government (MSIT) [RS-2021-II211341, Artificial Intelligence Graduate School Program (Chung-Ang University)] and a project for Collabo R\&D between Industry, University, and Research Institute funded by Korea Ministry of SMEs and Startups in 2025.(RS-2025-02323112).
\bibliography{acl_latex}
\appendix
\newpage
\clearpage
\section{Appendix}



\section{Prompts for SAFE-SQL}
\subsection{Prompt for example generation.}
\label{appen:example_prompt}
For example generation, we use zero shot prompt as shown in the figure~\ref{tab:example_generation}. 
\begin{table}[h]
\centering
\scalebox{0.85}{
\begin{tabular}{l}
\hline
\begin{tabular}[c]{@{}p{1.0\linewidth}@{}}
You are a powerful text-to-SQL reasoner. Your task is to generate ten similar questions, ten SQL queries, and ten reasoning paths for how the SQL queries are derived.
To ensure high-quality examples, focus on the following three key aspects:\\ \\ 
\textbf{Semantic Similarity}\\
Ensure that all generated questions have the same underlying meaning as the test question. Variations in wording, synonyms, and phrasing are allowed as long as they preserve the intended query objective.
Avoid introducing ambiguity or additional constraints that alter the intent.\\ \\ 
\textbf{Structural Similarity}\\
While key terms (such as table names, column names, and numerical values) may vary, their functional roles and relationships should remain intact. \\ \\
\textbf{Reasoning Path}\\
The logical reasoning required to construct the SQL query should remain consistent across examples. Clearly outline each step, including how key conditions are identified and mapped to SQL operations.Maintain coherence in how joins, aggregations, filters, and sorting operations are applied. 

Do not explain me about the result and just give me ten examples.
\\ \\
\textbf{\#\# Schema linking:} {schema\_linking[i]} \\
\textbf{\#\# Tables:} {test\_table[i]} \\
\textbf{\#\# Foreign keys:} {test\_foreign\_keys[i]} \\
\textbf{\#\# Question:} {test\_question[i]} \\ \\
\textbf{\#\# Similar Question:} \\
\textbf{\#\# SQL query:} \\
\textbf{\#\# Reasoning Path:} \\
\end{tabular}  \\ 
\hline
\end{tabular}
}
\caption{The zero-shot prompt used for example generation}
\label{tab:example_generation}
\end{table}

\subsection{Prompt for filtering examples.}
\label{appen:filtering_examples}
For example generation, we use zero shot prompt as shown in figure~\ref{tab:filtering_exmaples}. 
\begin{table}[]
\centering
\small
\scalebox{0.95}{
\begin{tabular}{l}
\hline
\begin{tabular}[c]{@{}p{1.0\linewidth}@{}}
You are a powerful text-to-SQL reasoner. Given a test question and a set of examples, compute the relevance score for each example based on the following criteria. Do not explain me about the answer, just give me scores.  \\ \\
\textbf{\*\*Semantic Similarity} \\
Compare the overall meaning of the test question and the example question.
Higher scores should be assigned if the two questions have the same intent, even if they are phrased differently.
Consider synonyms, paraphrasing, and minor wording variations that do not alter the fundamental meaning.
Assign lower scores if the test and example questions focus on different database operations (e.g., aggregation vs. filtering) or require fundamentally different types of information.(up to 10 points).\\
10: Almost identical meaning and intent.\\
7–9: Minor paraphrasing but highly relevant.\\
4–6: Some overlap but different focus.\\
1–3: Mostly unrelated meaning.\\
0: Completely different intent.
\\ \\
\textbf{\*\* Structural Similarity} \\ 
Evaluate the structural alignment between the test question and the example question by analyzing how key elements (such as entities, attributes, and numerical values) are connected. Even if individual nouns, verbs, or numbers differ, the overall relational structure should be considered. Focus on whether the dependencies between key components (e.g., how entities relate to each other in the database) remain consistent.(up to 10 points). \\
10: Nearly identical structural relationships and dependencies. \\
7–9: Mostly similar structure, with minor differences in entity connections. \\
4–6: Some overlap, but noticeable differences in how key components interact. \\
1–3: Few shared structural relationships, making alignment weak. \\
0: No meaningful structural similarities. 
\\ \\ 
\textbf{\*\*Reasoning Path} \\ 
Evaluate whether the logical steps needed to answer the example question align with those required for the test question. Consider whether the database operations (e.g., filtering, aggregation, joins, subqueries) are similar.A high score should be given if the example follows the same logical sequence to derive the SQL query.Lower scores should be assigned if the reasoning process differs significantly, even if the questions seem similar at a surface level.(up to 10  points). \\
10: Exact reasoning process to get right SQL query.\\
7–9: Mostly similar but with minor differences.\\
4–6: Some alignment but different key steps.\\
1–3: Largely different reasoning.\\
0: Completely unrelated logic. \\ \\

\textbf{\#\# Question:} {test\_question[i]} \\ 
\textbf{\#\# Similar Question:} {similar\_question[i]} \\
\textbf{\#\# Reasoning Path:} {reasoning\_path[i]} \\
\textbf{\#\# Relevance score:} 
\end{tabular}  \\ 
\hline
\end{tabular} 
}
\caption{The zero-shot prompt used for filtering examples.}
\label{tab:filtering_exmaples}
\end{table}

\subsection{Prompt for final inference.}
\label{appen:final_inference}
For final inference, we use zero shot prompt as shown in figure~\ref{tab:final_prompt}. 
\begin{table}[]
\centering
\scalebox{0.85}{\begin{tabular}{l}
\hline
\begin{tabular}[c]{@{}p{1.0\linewidth}@{}}\\
You are a powerful text-to-SQL reasoner. Your task is to generate the final SQL query using a set of selected examples that provide guidance on query construction. Utilizing Selected Examples. Do not explain me about the answer, just give me SQL query. \\
A set of chosen examples, each containing:
A natural language question similar to the test question
A corresponding SQL query
A detailed reasoning path explaining how the SQL query was derived
These examples are selected based on three key criteria:
\\ \\ 
\textbf{Semantic Similarity}
The selected examples closely match the intent of the test question.
Variations in wording do not change the meaning.
\\
\textbf{Structural Similarity}
The database schema elements (tables, columns, joins) used in the examples align with the test question.
The SQL syntax and structure are relevant to the expected query.
\\
\textbf{Reasoning Path Similarity}
The logical steps used to construct the SQL query align with the reasoning required for the test question.
Key transformations, filtering conditions, and aggregation logic are similar.
\\
\textbf{Final SQL Query Construction} \\
Using the selected examples, generate the final SQL query that correctly retrieves the desired result for the given test question.
Follow the reasoning patterns observed in the examples.
Now, generate the final SQL query for the given test question: \\ \\ 
\textbf{\#\#Tables:} test\_table[i]\\
\textbf{\#\#Foreign\_keys:} test\_foreign\_keys[i]\\
\textbf{\#\#Question:} text\_question[i] \\
\textbf{\#\#Filtered\_example:} filtered\_example[i] \\

\end{tabular}  \\ \hline
\end{tabular}}
\caption{The zero-shot prompt used for Final SQL query inference.}
\label{tab:final_prompt}
\end{table}

\label{appen:Impact_size}

\section{Impact of model size}

\begin{table}[t]
\centering
\small
\begin{tabularx}{\linewidth}{lXXXXX}
\toprule
\textbf{}  & \textbf{Easy} & \textbf{Med} & \textbf{Hard} & \textbf{Extra} & \textbf{All} \\ \midrule
\textbf{Qwen 2.5-3B}  & 62.4          &     61.2        &  58.6         & 48.8          & 59.1         \\
\textbf{Qwen 2.5-7B}  & 80.0         & 78.0           & 67.2         &   51.8        & 72.3        \\
\textbf{Qwen 2.5-14B} & \textbf{81.2}          & \textbf{80.3}            & \textbf{69.5}         & \textbf{56.4}          & \textbf{74.7}      \\ 
\bottomrule
\end{tabularx}
\caption{Execution accuracy performance of different size of models of Qwen series across difficulty levels of spider dev set.}
\label{tab:qwen_ab_models}
\end{table}


\paragraph{Performance based on generated examples across different model size}
As shown in Table~\ref{tab:qwen_ab_models}, We investigate the impact of model size on example generation with different variants of the Qwen2.5 Models. The results demonstrate that the 14B model achieves the highest overall performance, followed by the 7B and the 3B. This trend is consistent across all difficulty levels, with large model size generating higher-quality examples that lead to more accurate SQL query generation. The performance improvement with increasing model size can be attributed to the enhanced capacity of larger models to capture SQL question patterns and semantic relationships. Moreover, larger models possess more extensive information, allowing them to generate more appropriate questions and construct detailed reasoning paths, which contribute to the overall accuracy of SQL query generation.

\section{Spider dev training set embedding clusters.}
\begin{figure}[h]
\centerline{\includegraphics[scale=0.35]{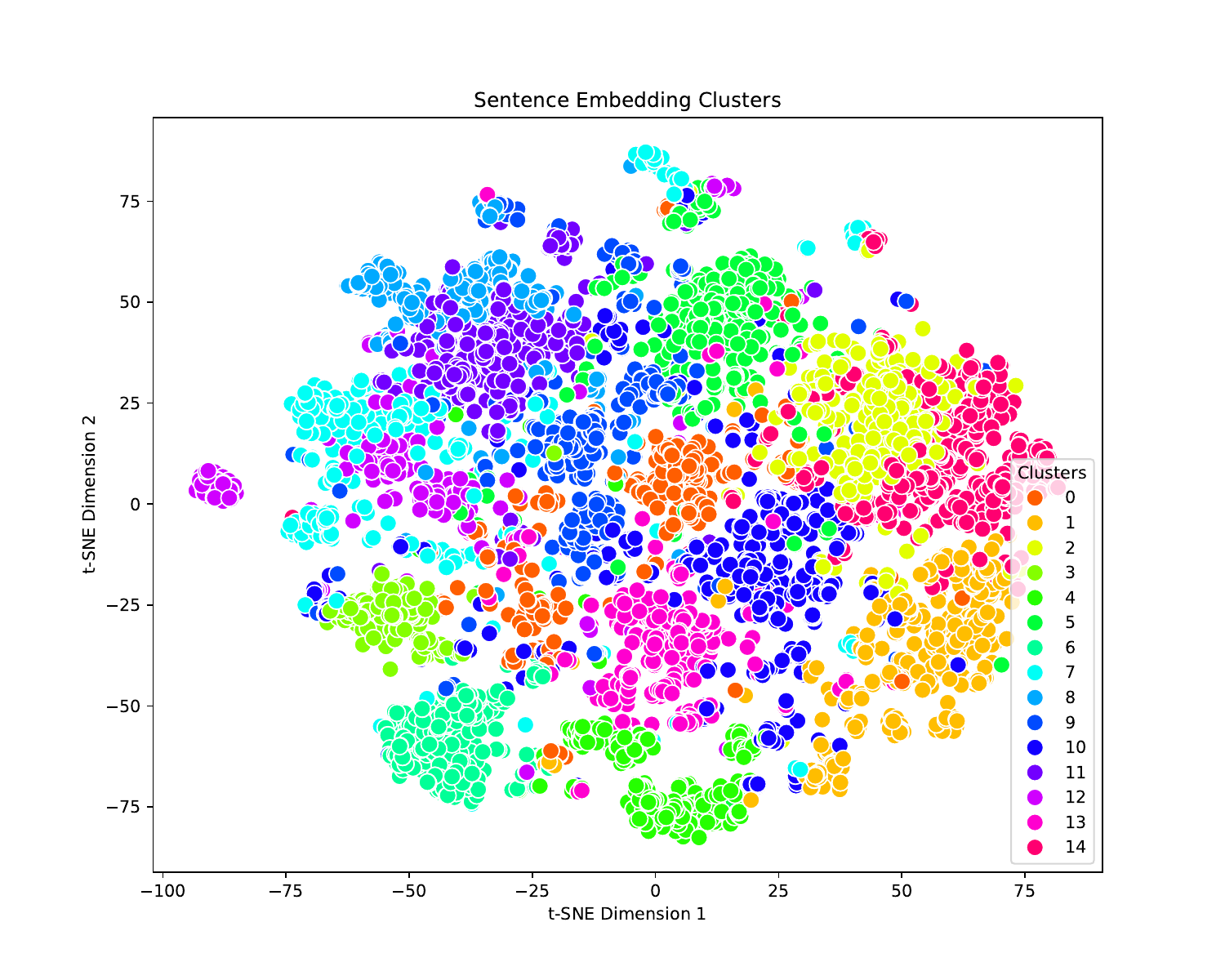}}
\caption{Embedding of spider dev set training questions.}
\vspace{-5mm}
\label{tab:spider_dev_embedding}
\end{figure}
 Although questions within the same category share semantic similarities, they may belong to different clusters, leading to inconsistencies when retrieving examples from the training set. This highlights the limitations of training set retrieval in Text-to-SQL tasks.





\end{document}